# MIAR: Modality Interaction and Alignment Representation Fuison for Multimodal Emotion Recognition

Jichao Zhu, Jun Yu*, *Member, IEEE*

***Abstract*—Multimodal Emotion Recognition (MER) aims to perceive human emotions through three modes: language, vision, and audio. Previous methods primarily focused on modal fusion without adequately addressing significant distributional differences among modalities or considering their varying contributions to the task. They also lacked robust generalization capabilities across diverse textual model features, thus limiting performance in multimodal scenarios. Therefore, we propose a novel approach called Modality Interaction and Alignment Representation (MIAR). This network integrates contextual features across different modalities using a feature interaction to generate feature tokens to represent global representations of this modality extracting information from other modalities. These four tokens represent global representations of how each modality extracts information from others. MIAR aligns different modalities using contrastive learning and normalization strategies. We conduct experiments on two benchmarks: CMU-MOSI and CMU-MOSEI datasets, experimental results demonstrate the MIAR outperforms state-of-the-art MER methods.**

***Index Terms*—Mulitmodal Emotion Recognition, Modality Fusion, Alignment.**

## I. Introduction

HUMAN emotions are not only revealed through the intonations of voice and facial expressions but also subtly influence physiological indicators such as brainwaves, muscle currents, and skin temperature. Ultimately, emotion recognition [1]–[4] is a comprehensive problem that involves multiple modalities. However, the difficulty in recognizing physiological signals and the challenges in data collection and pre-processing have limited the creation of current datasets. As a result, researchers typically use videos and their transcribed speech and text to study emotion recognition.

With the rapid development of artificial intelligence, breakthroughs have been continuously achieved in the fields of computer vision and speech recognition through Convolutional Neural Networks (CNNs) [5] and Visual Transformers [6]. In the field of natural language processing, large models such as ChatGPT have emerged. These outstanding research achievements have shown effective representations that often perform well in downstream tasks. This has drawn increasing attention to the field of emotion analysis. Studies have shown that unimodal emotion recognition applications are less effective than multimodal ones. This is because when there is a significant difference or lack in the collection and training dataset of one modality, other modalities can be used to reduce the impact of that modality.

Introducing feature engineering from the aforementioned fields into the multimodal emotion domain is an important and challenging task. Effective modality representation needs to comprehensively consider the homogeneity and heterogeneity between different modalities. Homogeneity refers to the consistent purpose conveyed by various modalities simultaneously. For example, when humans express sadness, it is usually accompanied by facial crying, sobbing sounds, key phrases, and electromyography, reflecting the common characteristics across different modalities. Heterogeneity refers to the individual attributes of each modality, such as voice tone and fixed gestures, which are unrelated to the purpose and are non-common characteristics reflected in the modality's features.

In the field of multimodal emotion recognition research, how to effectively extract representative features from various modalities has become a hot topic of great concern. Excellent feature engineering can not only reveal the deep-level information behind the data but also often demonstrates outstanding generalization capabilities in subsequent tasks. Addressing this issue, our research fully considers the following key points and proposes an innovative method based on this: We have noticed significant distributional differences between different modalities. The data characteristics of audio, visual, and text modalities are distinct from one another, and these differences have an important impact on the accuracy and efficiency of emotion recognition tasks. Therefore, effectively addressing these differences is key to achieving cross-modal emotion recognition. Secondly, the contribution of different modalities to the emotion recognition task is not the same. For example, in this task, textual information may play a decisive role. Our method aims to balance and optimize the feature representation of each modality to maximize its contribution to the overall task. We have also taken into account the generalization ability of a large number of text model feature representations. In multimodal scenarios, how to avoid excessive reliance on specific modalities and improve the model's generalization ability across different data distributions is a challenging issue. Based on the above discussion, we proposes a novel **M**odality **I**nteraction and **A**lignment **R**epresentation (MIAR) approach based on dual text encoders. Through the design of this dual encoder, the MIAR method can better handle the issues of inter-modality interaction and alignment, thereby achieving more accurate and robust results in cross-modal emotion

Jichao Zhu and Jun Yu are with the Department of Automation, University of Science and Technology of China, Hefei, Anhui 230022, China. E-mail: jichaozhu@mail.ustc.edu.cn; harryjun@ustc.edu.cn.
* is the corresponding author.

recognition tasks.

Specifically, this paper decomposes each modality into homogeneous and heterogeneous spaces. In the homogeneous subspace, modality homogeneous features are generated through encoders with shared weights. In the heterogeneous space, modality interaction is first performed, and then the text-text pair features are aligned through contrastive learning. The same text features are aggregated through the distribution of the two text encoder outputs, aligning audio-visual features through contrastive alignment, as well as text-audio and text-visual alignment methods, making audio and visual features closer in the projection space while retaining the generalization ability of the text.

The main contributions can be summarized as follows:

- This paper proposes a fusion framework for modality interaction, which is applicable to cross-modal emotion recognition tasks in scenarios where features have already been extracted;
- Two alignment strategies are adopted, namely contrastive learning alignment and norm alignment, which work together to obtain more effective representations for audio, visual, and text modalities;
- Extensive experiments have been conducted on the CMU-MOSI and CMU-MOSEI emotion analysis datasets, and the proposed method has achieved competitive results across various experimental metrics.

## II. Related Works

In the method of contrastive learning within samples, Yang et al. [7] proposed a unified learning framework called Contrastive Feature Decomposition (ConFEDE), which simultaneously performs contrastive representation learning and contrastive feature decomposition within the features of the samples. Zhao et al. [8] addressed language-text pairs by proposing a two-stage method of contrastive pre-training (CLAP). In the first step, CLAP conducts contrastive pre-training on large-scale unlabeled data sets to enhance the representational capabilities of single modalities. In the second step, utilizing a multi-modal fusion architecture based on Transformers, they further refine and optimize the features of each modality through task-specific training, achieving sentiment classification. They highlighted the role of pre-training, contrastive learning, and representation learning in sentiment recognition tasks. Ma et al. [9] believe that the key terms before and after the text can lead to inaccuracies in sentiment analysis. To address this issue, they introduced PriSA, which resolves the issue of false relevance introduced by the integration of textual modalities and sentiment labels. It integrates priority fusion and distance-aware contrastive learning, proposing a prioritized inter-modal fusion method that uses textual modalities to guide the computation of inter-modal correlations, as well as a distance-aware contrastive learning method that utilizes inter-modal features to calculate mixed-modal correlations, obtaining more flexible multi-modal representations. Unlike traditional research, which mainly focuses on integration strategies between different modalities within sample pairs, Liu et al. [10] explored inter-reference relationships between samples and proposed a three-stage method incorporating multi-view contrastive learning to refine the accuracy of target representations. First, they used supervised contrastive learning to encourage the aggregation of samples of the same category on each modality while dispersing samples of different categories, strengthening the connections within the same category and distinguishing between different categories. Then, through an interaction module, they obtained more refined cross-modal representations and further refined these representations using self-supervised contrastive learning. Wu et al. [11] proposed a multi-modal hint gate module that converts non-verbal information into a multi-modal hint integrating textual context, filtering non-verbal signals with textual information, and an instance-to-label contrastive learning method, distinguishing different labels in the latent space semantically. These methods handle distribution differences between different modalities and integrate multi-source knowledge. Liu et al. [12] applied contrastive learning to modality-invariant features, reconstructing information of missing modalities using existing modality features and combining modality-invariant features with the reconstructed missing features for sentiment recognition. Addressing the subtle and transient characteristics of micro-expressions, Wang et al. [13] proposed an adaptive time-enhanced momentum contrastive learning method to mitigate these issues. For small-scale data, we used momentum contrast for contrastive learning and pre-trained the model on a new interpolated data set. To address the subtle and rapid facial movements, we further enhanced the temporal dynamics of redundant frames through an adaptive discard operation, improving the accuracy of expression recognition. Besides, many research are based on graph. Lin et al. [14] proposed a novel hierarchical graph contrastive learning (HGraph-CL) framework to understand complex intra- and inter-modal relationships, constructing separate unimodal graphs for each modality to capture modality-specific emotional meanings. By adopting a graph learning strategy, they explored potential relationships enhanced by unimodal graphs, and then integrated the unimodal graphs to create a multimodal graph capable of understanding graph structures used for learning complex relationships between different modalities. Zhang et al. [15] proposed a hierarchical multimodal fusion graph neural network (HMF-GNN) for real-world health multimodal datasets, aggregating long-term and multimodal information (e.g., medication and treatment) from heterogeneous neighbor nodes. They represented multimodal information and longitudinal records of human behavior as a dynamic heterogeneous graph, obtaining accurate node embeddings on this graph to better learn graph topological information. Tang et al. [16] proposed a dual graph structure network (DGS-Net) for emotion recognition in multimodal dialogues. This network uses a dual graph structure-based multimodal fusion mechanism to represent multimodal information. Specifically, it extracts heterogeneous features of each modality through separate graph structures and explores complementary features between modalities through aggregated graph structures, achieving full fusion of modality information. Additionally, a local attention mechanism was designed to limit the scope and target of emotion analysis. Tu et al. [17] developed an adaptive interaction graph network model called

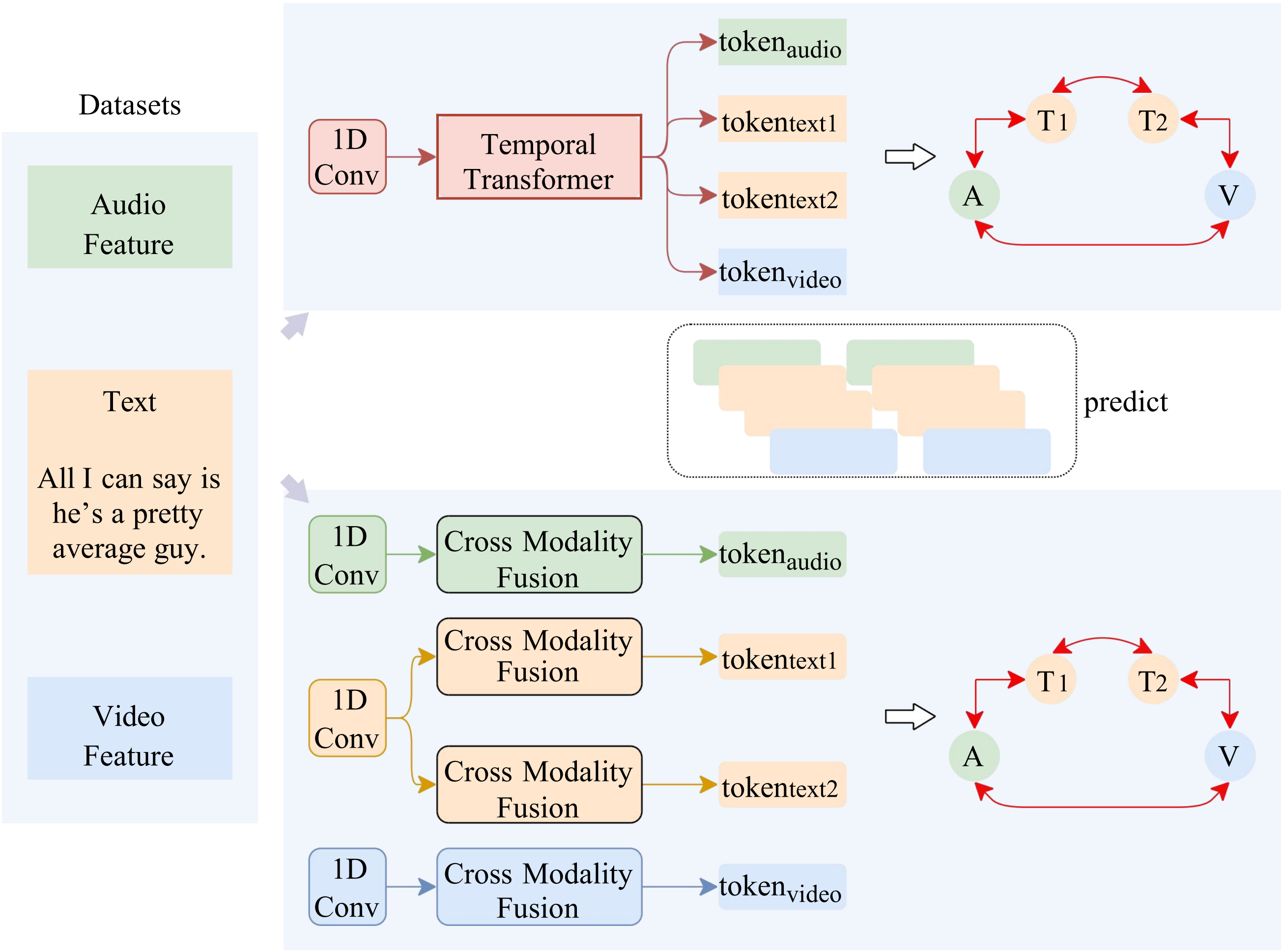

Fig. 1. Modality Interaction and Alignment Representation Fusion Prediction Framework (MIAR)

AdaIGN, which enhances intra- and cross-modal interactions by adaptively selecting nodes and edges, while using directed IGN to prevent future utterances from affecting current ones, achieving multimodal emotion recognition. Li et al. [18] proposed the DMD method, which decouples according to modality heterogeneity and adapts the transmission of domain knowledge by treating each modality as a node and providing graph distillation units for each decoupled part, based on the different contributions of modalities to the emotion recognition task.

## III. Methodology

This section comprises four parts: the algorithm framework diagram, text feature extraction, modality interaction fusion, and cross-modal alignment strategy.

### A. *Framework Diagram*

The emotion recognition framework for multimodal alignment representation fusion proposed in this chapter is shown in Figure 1. The entire recognition process can be divided into four parts: existing audio and visual features and text features to be extracted as model input data, cross-modal interaction fusion, cross-modal alignment, and emotion prediction.

Firstly, the input data for the model comes from the dataset, including existing features and corresponding textual sentences. These raw texts are used as inputs for two different text encoders to generate a set of features for the samples. Since the feature dimensions extracted by different models are not entirely consistent, a one-dimensional convolution mapping capability is used to unify the different feature dimensions, ensuring that the features from the three modalities, totaling four, can serve as inputs for the cross-modal interaction fusion module. Here, there are four one-dimensional convolution modules that map the three modalities into homogeneous and heterogeneous subspaces. The convolution and sequence modules of the former share weights.

After fusion, modality-specific tokens are generated, including audio tokens, visual tokens, and two text tokens. The reason for having two text tokens is due to the presence of two text encoders, with each token representing the result of feature interaction with other modalities. Next, these four tokens are used as inputs for the cross-modal alignment module. Through contrastive learning between text-text and audio-visual pairs, the spatial distribution of the corresponding features is altered, making the distribution within the same sample closer. For audio-text and visual-text pairs, norm alignment is used to align the two modalities.

This process aims to ensure consistency and correlation between modalities, so that the modality representations can be better utilized in subsequent tasks. Through the above alignment strategy, the model can effectively leverage information from different modalities and establish effective links between them, thereby improving overall performance and generalization ability.

### B. Text Encoder

Large models pre-trained using self-supervised methods have achieved continuous breakthroughs in natural language processing and other fields, making text feature engineering more effective than audio and video. In this paper, we will use open-source large language models for text feature extraction, including CLAP, Chat-GLM, and others, and further utilize them as text features for experiments. For this purpose, we leverage the HuggingFace open-source platform, which provides convenient interfaces to call the aforementioned large models. It is worth noting that many of these large language models are based on the Transformer architecture. The training data includes inputs from different modalities, and contrastive loss functions are incorporated during the training process. The text encoders of these models perform excellently when handling text and cross-modal tasks. Therefore, we choose to use these models to extract text features to build our experimental system. When processing text data, it is usually necessary to convert raw text into a form that machines can understand and process. Generating text features is a common method that converts textual information into vector representations, allowing the model to process and analyze them.

In this process, tokenization is performed first. Tokenization is the process of dividing text into individual tokens or subwords. After processing by the tokenizer, the original sequence is mapped into a discrete numerical sequence $T = \{t_1, \cdots, t_n\}$ that the model can handle. For example, in Chinese text, a sentence can be divided into individual characters or finer-grained subword units. The purpose of this is to break the text into discrete numbers for subsequent processing.

Next, feature extraction is performed through the pre-trained text encoder. The discrete sequence $T$ is mapped to a continuous vector space through an embedding layer to better represent the semantic relationships between features. This helps the model better understand the meaning and structure of the text. In this process, each token or subword is mapped to a feature vector. The first feature vector often corresponds to a special token, such as the CLS token, which can be used to represent the entire text sequence in subsequent processing.

The continuous feature sequence is input into the text encoder, and through the computations of these layers, the model can understand and model the text at the feature level and contextual level, generating features $F = \{CLS, f_1, \cdots, f_n\}$.

Overall, the process of generating text features involves tokenization, feature extraction, embedding representation, and the use of text encoders, as well as further processing of the output features. This approach helps the model better handle and understand text data.

### C. Modality Interaction Fusion Module

*1) Temporal Feature Augmentation Module:* The individual audio and visual features provided in the dataset come from time windows and each frame, respectively. However, multimodal emotion recognition targets continuously changing content, with inputs being temporally dimensioned samples. Therefore, it is necessary to consider the temporal sequence. The temporal feature module is mainly used for the features of text, audio, and visual modalities. The raw modalities are represented as $I_t$, $I_a$, and $I_v$, respectively. They are converted to the same dimension through one-dimensional convolution with an output channel of 50, denoted as $d_{model}$=50, and represented as:

$$F_m = \text{Conv1D}(I_m), \quad m \in \{t, a, v\} \tag{1}$$

where $F_m$ represents the features after the mapping transformation, and $m$ denotes the modality.

The Temporal Feature Augmentation Module (TFA), effectively utilizes the contextual information within different modalities of the sample to extract common features. Starting from the self-attention mechanism, the local features of the three modalities interact with the global context information within the modality to model the global temporal sequence and obtain the global temporal dependencies. The modality tokens are then derived based on the generated temporal features.

Taking the visual modality features $F_v \in R^{B \times T \times d}$ as an example, where $T$ represents the number of frames in the input video, the features are transformed into query $Q$, key $K$, and value $V$ through three fully connected layers. The attention score matrix $S$ is generated through scaled dot-product operations and softmax normalization, with a shape of $R^{T \times T}$. Here, $s_{ij}$ represents the attention score between the $i$-th and $j$-th frames. This matrix is multiplied with the value matrix $V$ to generate new features containing weighted temporal information associations.

*2) Cross-Modality Interaction Module:* Considering that different modalities represent the same labels, cross-modality fusion of diverse and complementary modalities can provide more useful information. This section describes the principles of a cross-modality fusion network designed to handle feature interaction and contextual relevance between three different modalities.

Firstly, the network employs a cross-attention mechanism, meaning features from different modalities influence and communicate with each other. Taking the text feature $F_t \in R^{N \times L \times d}$ as an example, it interacts with audio features $F_a \in R^{N \times L \times d}$ and video features $F_v \in R^{N \times L \times d}$, where N is the batch size, L is the feature length, and d is the dimension after one-dimensional convolution mapping, i.e., $d_{model}$. This interaction helps to fuse information from different modalities to capture their correlations and contextual information. To achieve cross-modality fusion, this paper proposes a Cross Modality Transformer (CMF) network structure, as shown in Figure 2. This network fuses the input text, audio, and visual features $F_t$, $F_a$, and $F_v$ using a cross-attention mechanism and outputs the fused features through a temporal module. These projected features will serve as the input for subsequent parts. Note that the overall framework determines the number of fusion networks based on the number of input modality features. In this study, there are four groups of input modality features, thus there will be four CMF networks. By utilizing the cross-modality feature fusion module, it effectively captures the correlations between different modalities, generating feature tokens, thereby improving the model's performance in multimodal data processing tasks.

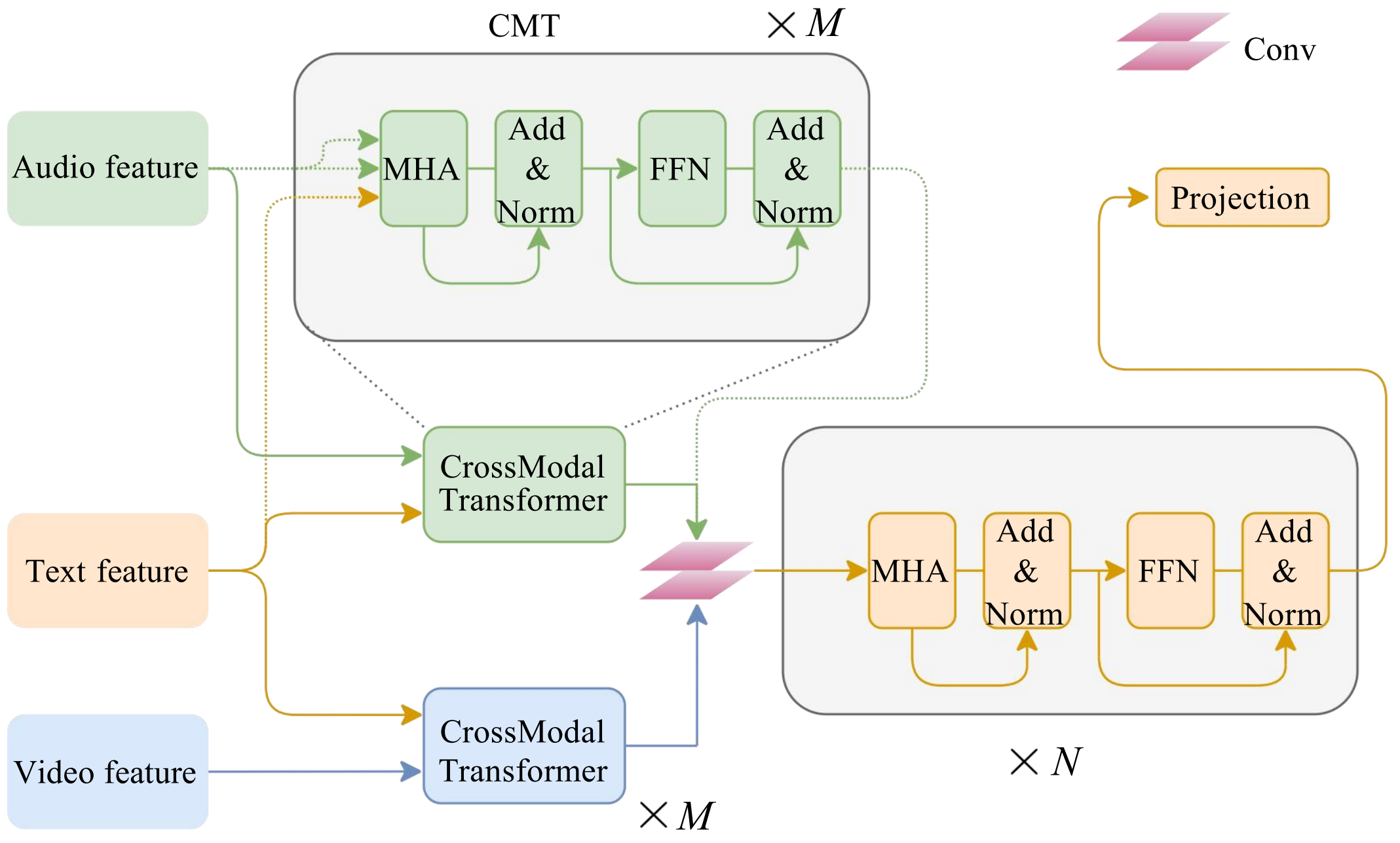

Fig. 2. Cross-Modality Feature Fusion Module (using text features as an example)

As part of the model input, audio features and visual features are separately fed into two Cross Modality Transformer (CMT) networks. In each CMT, K and V come from $F_a$ or $F_v$, with text features $F_t$ serving as the query $Q_t$. When K is audio, $Q_t = F_t W_{Q_t}$, $K_a = F_a W_{K_a}$, $V_a = F_a W_{V_a}$, where the learnable parameters $W_{Q_t} \in R^{d \times d/n_h}$, $W_{K_a} \in R^{d \times d/n_h}$, and $W_{V_a} \in R^{d \times d/n_h}$, and the hyperparameter $n_h$ is the number of heads in multi-head attention, which can map to subspaces representing different points of focus, thereby capturing and processing information from various points of focus simultaneously. Through the Cross-Modality Multi-Head Attention layer (MHA), the features of each subspace are concatenated and passed through a linear layer to obtain the output of the multi-head attention layer.

The output of cross-attention is given by equation 2:

$$\begin{aligned}\text{Attention}(Q_t, K_a, V_a) &= \text{softmax}\left(\frac{Q_t {K_a}^T}{\sqrt{d}}\right) V_a \\ &= \text{softmax}\left(\frac{F_t W_{Q_t} (F_a W_{K_a})^T}{\sqrt{d}}\right) F_a W_{F_a}\end{aligned} \tag{2}$$

The $(i, j)$ entry of this matrix represents the attention of the i-th text character to the j-th time window of the audio feature. Then, through the Feed-Forward Network (FFN) and residual layer (Residual), it generates the fused features $F_{ta}$ and $F_{tv}$, as shown in equation 3:

$$\begin{aligned} F_{ta} &= \text{CMT}(F_t, F_a) \\ F_{tv} &= \text{CMT}(F_t, F_v) \end{aligned} \tag{3}$$

Next, after stacking M layers for thorough bimodal interaction, the fused features $F_a$ and $F_v$ are extracted.

To adaptively weight the fusion of the above two interaction features, it is necessary to introduce the channel reconfiguration capability, dimensionality reduction, and computational efficiency of $1 \times 1$ convolution. Therefore, this paper uses $1 \times 1$ convolution to fuse the text feature-extracted audio modality features $F_a$ and visual modality features $F_v$. In this method, the input channel number of the two-dimensional convolution is 2, and the output channel number is 1, expressed as:

$$F_{t,\ av} = \text{Conv2D}(F_{ta},\ F_{tv}) \tag{4}$$

Compared to direct dimensional concatenation, it reduces spatial dimensions and computation time for subsequent temporal modules, while the $1 \times 1$ convolution has the ability to mix and weight the channel information of the two.

Next, the merged $F_{t,av}$ is input into the temporal feature enhancement module based on the self-attention mechanism to aggregate temporal information, represented as:

$$F^{'}_{t,\ av} = \text{TFA}(F_{t,av}) \tag{5}$$

Since the models extracting audio and visual features do not have a clear identifier like the CLS token in large text models, the vector at the 0th position of the sufficiently fused feature $F^{'}_{t,av}$ after N layers already contains important information of the entire sequence. Therefore, this chapter uniformly treats this vector as a token representing the entire sequence feature. In this way, we obtain a vector whose length is consistent with the input sequence length. Then, after processing by the projection layer, a text token is generated from the feature fusion network, i.e., a text feature corresponds to a feature token, represented as:

$$\text{text-token} = F^{'}_{t,av}[0] \tag{6}$$

Similarly, when audio features serve as the query $Q$, the text features from the two text encoders are averaged element-wise and used as the text input features for the audio modality CMT, as follows:

$$\begin{aligned} F_{at} &= \text{CMT}(F_a, \frac{F_t^1 + F_t^2}{2}) \\ F_{av} &= \text{CMT}(F_a, F_v) \end{aligned} \tag{7}$$

When video features serve as the query $Q$, similarly:

$$\begin{aligned} F_{va} &= \text{CMT}(F_v, F_a) \\ F_{vt} &= \text{CMT}(F_v, \frac{F_t^1 + F_t^2}{2}) \end{aligned} \tag{8}$$

The features in equations 7 and 8 are concatenated and used as the input for the TFA module, generating audio-token and video-token, as shown in equation 9:

$$\begin{aligned} F_{a,\ tv} &= \text{Conv2D}(F_{at},\ F_{av}) \\ F_{v,\ ta} &= \text{Conv2D}(F_{vt},\ F_{va}) \\ \text{audio-token} &= F^{'}_{a,\ tv}[0] \\ \text{video-token} &= F^{'}_{v,\ ta}[0] \end{aligned} \tag{9}$$

Thus, after processing through four CMF modules, the audio-token, text-token$_1$, text-token$_2$, and video-token are generated, which will serve as the input for the next module in the following section.

### *D. Alignment Module*

In this subsection, we explore the alignment of speech, visual, and text modalities. In the field of multimodal emotion recognition, text-based methods perform best. However, considering the differences between modalities, we introduce two text encoders, each including contrastive learning with speech and visual modalities during training. This reduces the gap and distribution differences between different modality features. Therefore, it is necessary to align the combinations of four modalities: text-text, speech-visual, speech-text, and visual-text. These alignment operations aim to ensure consistency between different modalities and improve the model's performance.

When processing features from different modalities, the semantic spaces may not be shared, leading to differences in feature spaces. To address this, we align these modality feature representations in a shared semantic space to enhance their comparability and relevance. Firstly, we use multilayer perceptrons (MLP) to transform feature representations from different spaces. MLP is a common neural network structure that maps input data to different dimensional spaces. By applying MLP transformation to labels from different modalities, we map them to the same feature space. Higher-dimensional representations can be richer but more complex, while lower-dimensional mappings may help the model generalize better to unseen data by learning more general features and patterns rather than memorizing training data.

Figure 3 illustrates the proposed cross-modality alignment strategy, which includes two components: a contrastive alignment matrix, where elements represent the similarity between corresponding samples, and norm alignment.

*1) Contrastive Alignment Strategy:* Using text as an example, we explain our contrastive alignment strategy, which uses contrastive learning methods to optimize the alignment of text labels. Contrastive learning is an unsupervised learning method that maximizes the similarity between positive samples (samples with the same semantics) and minimizes the similarity between negative samples (samples with different semantics) to learn feature representations.

In this process, this chapter defines different modalities of the same sample as positive samples, constructing a similarity matrix by calculating the similarity between them. Suppose $t_i^1$ and $t_j^2$ represent the text labels corresponding to the first and second encoders, respectively, and $i$ and $j$ represent the positions of samples in each batch. The similarity measure used for the contrastive alignment similarity matrix M is cosine similarity, where a value closer to 1 indicates higher similarity, and vice versa. Therefore, in Figure 3, each element $m_{ij}$ of M is given by equation 10:

$$\begin{aligned} \text{sim}(t_i^1, t_j^2) &= \cos(t_i^1, t_j^2) = \frac{t_i^1}{\|t_i^1\|_2} \cdot \frac{t_j^2}{\|t_j^2\|_2} \\ m_{ij} &= \text{sim}(t_i^1, t_j^2) \end{aligned} \tag{10}$$

Thus, the contrastive loss terms for the two text labels are given by:

$$\begin{aligned} \text{softmax}(m_{ij}) &= \frac{e^{m_{ij}\ /\ \tau}}{\sum_{j=1}^{N} e^{m_{ij}\ /\ \tau}} \\ \mathcal{L}_i^{12} &= -\log \frac{\exp(\text{sim}(t_i^1,\ t_i^2)\ /\tau)}{\sum_{j=1}^{N} \exp(\text{sim}(t_i^1,\ t_j^2)\ /\ \tau)} \\ \mathcal{L}_j^{21} &= -\log \frac{\exp(\text{sim}(t_j^2,\ t_j^1)\ /\ \tau)}{\sum_{i=1}^{N} \exp(\text{sim}(t_i^2,\ t_j^1)\ /\ \tau)} \end{aligned} \tag{11}$$

where $N$ denotes the number of samples in each batch, and $\tau$ is a temperature hyperparameter that adjusts the distribution of the similarity matrix, typically set to 0.07. During loss computation, the rows and columns of $M$ undergo softmax normalization. Here, $L_i^{12}$ represents the cross-loss between the text token from the first encoder output and all text tokens from the second encoder output at position $i$, while $L_j^{21}$ represents the cross-loss between the text token from the second encoder output and all text tokens from the first encoder output at position $j$.

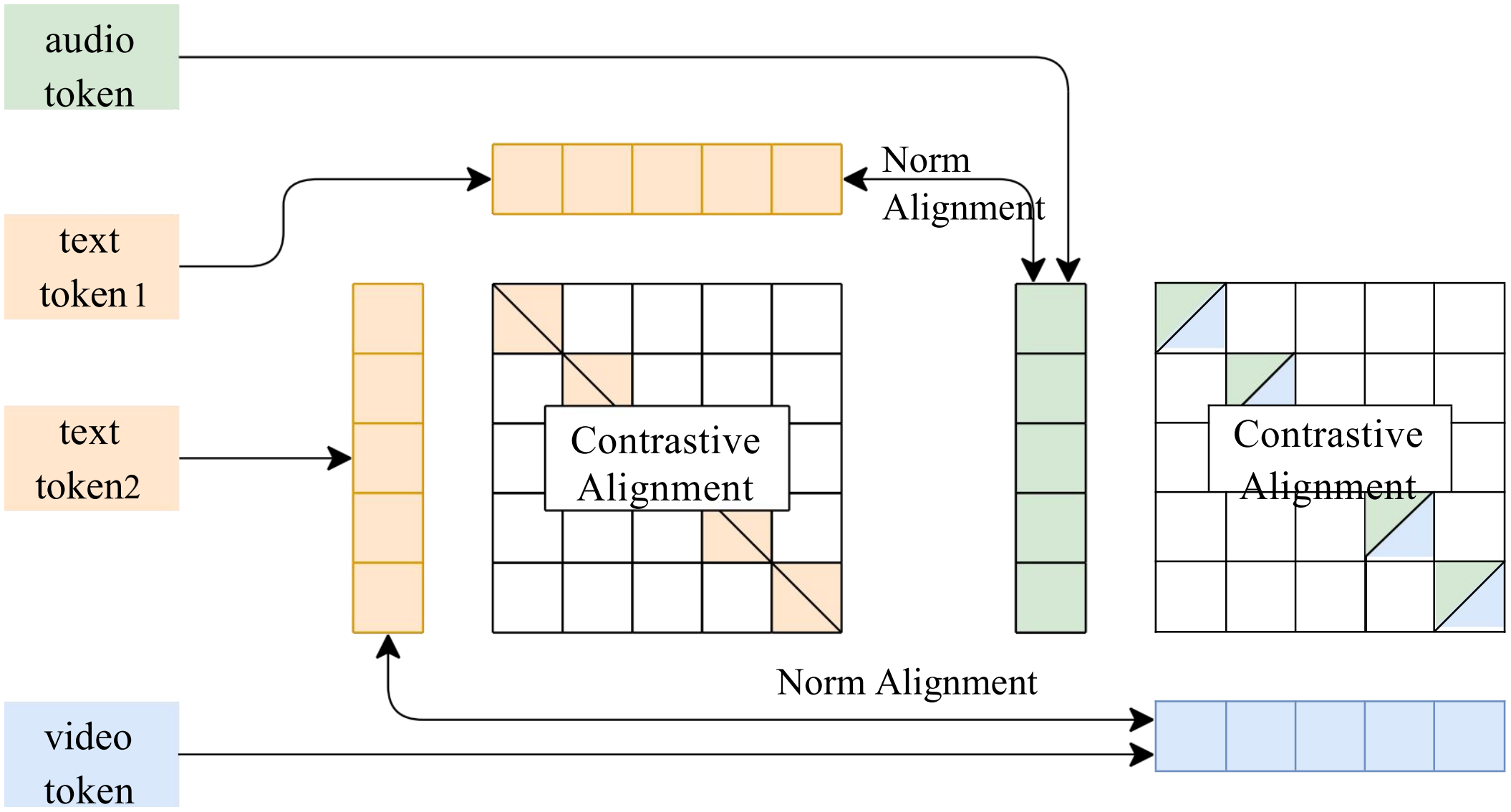

Fig. 3. Cross-Modality Alignment Strategy

Since the same text is defined as a positive sample, the text feature subscripts in the numerators of equation 11 are the same. In the matrix $M$, the diagonal positions are the positive samples, with a total of $N$ positive samples. The contrastive loss between the two text encoders is given by:

$$\mathcal{L}_{ttcl} = \frac{1}{2N}(\sum_{i=j}^{N}\mathcal{L}_i^{12} + \mathcal{L}_j^{21}) \tag{12}$$

For the speech and visual modalities, contrastive learning is also used to improve their feature space distribution. Let $a_i$ and $v_i$ denote the audio and video features, respectively. Similarly, the contrastive learning loss between audio and video can be expressed by equation 13:

$$\mathcal{L}_{avcl} = -\frac{1}{2N}\sum_{i=1}^{N}( \log \frac{\exp(\text{sim}(a_i,\ v_i)\ /\ \tau)}{\sum_{j=1}^{N}\exp(\text{sim}(a_i,\ v_j)\ /\ \tau)} + \log \frac{\exp(\text{sim}(v_i,\ a_i)\ /\ \tau)}{\sum_{j=1}^{N}\exp(\text{sim}(v_i,\ a_j)\ /\ \tau)}) \tag{13}$$

*2) Norm Alignment Strategy:* Contrastive learning for text-text and audio-visual aligns their internal distributions, but since text-based emotion recognition often outperforms video and audio, and the text encoder partially understands other modalities, norm alignment is used for text-audio and text-video alignment to reduce differences:

$$\mathcal{L}_{tatvm} = \frac{1}{2N}\sum_{i=1}^{N}(\|t_i^1 - a_i\|_p + \|t_i^2 - v_i\|_p) \tag{14}$$

where $p \in \{1, 2\}$, representing the L1 norm and L2 norm.

*3) Alignment Loss:* In summary, the alignment loss function includes contrastive loss (equations 12 and 13) and norm loss (equation 14), as shown in equation 15:

$$L_{align} = L_{ttcl} + L_{avcl} + \alpha L_{tatvm} \tag{15}$$

where $\alpha$ is a hyperparameter used to adjust the relative importance of these loss terms, ensuring that features from different modalities are aligned in a shared semantic space, thereby enhancing the model's generalization ability.

## IV. Experimental Setup and Analysis

### A. Datasets

The experiments in this chapter are based on several commonly used datasets in the field of emotion recognition, namely CMU-MOSI [19] and CMU-MOSEI [20], both of which are used for sentiment analysis with similar annotations.

The CMU-MOSI dataset serves as a widely used benchmark for evaluating the performance of fusion networks in emotion intensity prediction tasks. This dataset consists of monologues from numerous YouTube video blogs expressing opinions on specific topics. It comprises 2,199 video segments from 93 different videos presented by 89 different narrators. Each segment is manually annotated with a real-valued score ranging from $-3$ to $+3$, indicating the intensity of negative or positive emotions. Figure 4 shows the distribution of seven categories in the dataset, predominantly ranging from $[-2, 2]$. The CMU-MOSEI dataset was released in 2018 and is a multimodal affective domain dataset collected from YouTube. It enriches the versatility of speakers and covers a wider range of topics. The dataset contains 22,852 video slices, which are annotated continuously in the range of $[-3, +3]$, ranging from strong negative to strong positive: strong negative (-3), negative (-2), weak negative (-1), neutral (0), weak positive (+1), positive (+2), and strong positive (+3). These slices are taken from 5,000 videos, involving 1,000 different speakers and 250 different topics. In Figure 5, the proportion of each category in the dataset is displayed, where it can be seen that the neutral (0) category accounts for as high as 41.8%.

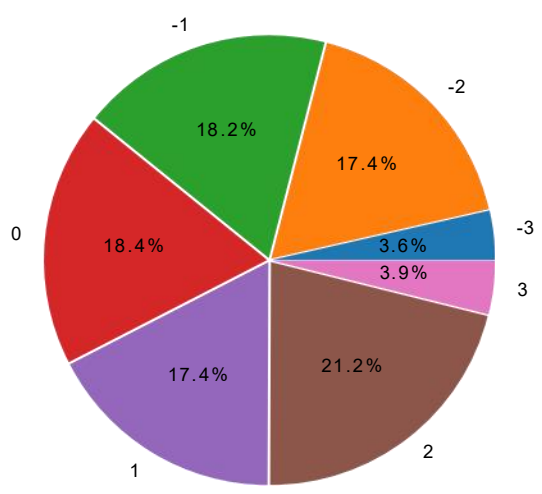

Fig. 4. CMU-MOSI's label distribution

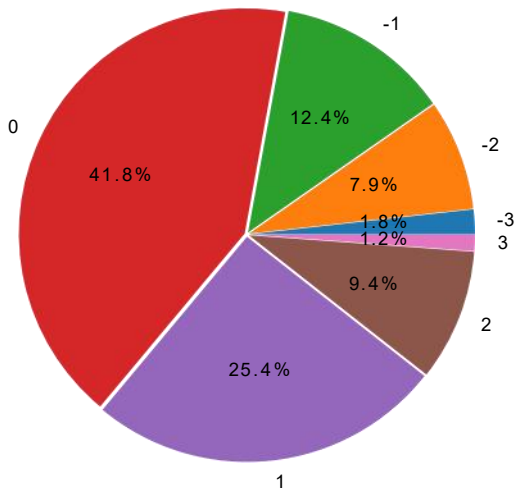

Fig. 5. CMU-MOSEI's label distribution

The original partitioning of samples in the CMU-MOSI and CMU-MOSEI datasets is shown in Table I.

TABLE I
SAMPLE COUNTS IN DATASETS

| Dataset | Training | Validation | Test |
|---|---|---|---|
| CMU-MOSI | 1284 | 229 | 686 |
| CMU-MOSEI | 16326 | 1871 | 4659 |

### B. Optimization Function

This paper integrates the aforementioned loss terms, and the combined optimization function is formulated as shown in Equation 16:

$$\begin{aligned} L_{total} &= L_{task} + W L_{align} \\ &= \mathrm{MSE}(y,\ \hat{y}) + \omega \mathcal{L}_{align} \end{aligned} \tag{16}$$

Here, $L_{task}$ represents the loss associated with the regression task. In the CMU-MOSI and CMU-MOSEI datasets, emotion annotations range continuously between $[-3, 3]$, hence the loss function used here is Mean Square Error (MSE). $L_{align}$ is the alignment loss combined as in Equation 15, where the importance of different constraints is controlled by adjusting the hyperparameter $W$.

### C. Evaluation Metrics

To evaluate MIAR against previous methods, we use common evaluation metrics including seven-class accuracy (Acc7), binary accuracy (Acc2), and the F1 score.

### D. Implementation Details

For ease of experimentation, the temporal length for audio, text, and video modalities is uniformly set to 50. Pre-extracted features are utilized for audio and video data from the dataset, with dimensions of 74 and 35 respectively. Text features are extracted using various pretrained models, maintaining their original dimensions. All input features are mapped to a unified dimension $d_{model} = 50$. The balance parameter for loss is set to 0.1. We use the PyTorch [21] to update model parameters on an NVIDIA RTX 3090 with 24GB, the optimizer used to optimize model parameters is Adam [22] with adaptive momentum, with a learning rate set to 0.0001. Batch sizes for datasets CMU-MOSI and CMU-MOSEI are set to 16 and 128 respectively. The training is conducted for 100 epochs with a seed of 101 for reproducibility.

### E. Results and Analysis

TABLE II
COMPARISON OF THE PERFORMANCE ON THE CMU-MOSI

| Methods | Acc2 (%, ↑) | F1 (%, ↑) | Acc7 (%, ↑) |
|---|---|---|---|
| IMDer | 85.7 | 85.6 | 45.3 |
| DiCMoR | 85.7 | 85.6 | 45.3 |
| GCNet | 85.2 | 85.1 | 44.9 |
| FDMER | 84.6 | 84.7 | 44.1 |
| MCTN | 79.3 | 79.1 | - |
| GraphCAGE | 82.1 | 82.1 | 35.4 |
| MIAR(our) | 86.59 | 86.57 | 40.67 |

TABLE III
COMPARISON OF THE PERFORMANCE ON THE CMU-MOSEI

| Methods | Acc2 (%, ↑) | F1 (%, ↑) | Acc7 (%, ↑) |
|---|---|---|---|
| IMDer | 85.1 | 85.1 | 53.4 |
| DiCMoR | 85.1 | 85.1 | 53.4 |
| GCNet | 85.2 | 85.1 | 51.5 |
| FDMER | 86.1 | 85.8 | 54.1 |
| MCTN | 79.8 | 80.6 | 49.6 |
| GraphCAGE | 81.7 | 81.8 | 48.9 |
| CORECT | 83.7 | - | 46.3 |
| COGMEN | 85.0 | - | 44.3 |
| MIAR(our) | 86.27 | 86.23 | 53.79 |

*1) Comparison on CMU-MOSI and CMU-MOSEI dataset:* This section compares the proposed method with several recent multimodal sentiment analysis approaches. These mainstream methods include those based on multimodal diffusion, multimodal probability distributions, graph neural networks, and others, detailed as follows:

1) IMDer [23] (Incomplete Multimodality-Diffused Emotion Recognition): Utilizes multimodal diffusion to map noise to the distribution space of missing data and reconstruct it.
2) DiCMoR [24] (Distribution-Consistent Modal Recovering): Introduced by Wang et al., this model maintains distribution consistency by transferring the distribution of existing modalities to other missing parts.

TABLE IV
THE RESULTS OF THE LANGUAGE MODEL ON A SINGLE MODALITY.

| Method | CMU-MOSI | | | CMU-MOSEI | | |
|---|---|---|---|---|---|---|
| | Acc2 (%) | F1 (%) | Acc7 (%) | Acc2 (%) | F1 (%) | Acc7 (%) |
| BERT | 84.12 | 84.10 | 44.34 | 84.17 | 84.20 | 52.60 |
| CLAP | 84.22 | 84.19 | 44.6 | 84.06 | 84.14 | 52.62 |
| ChatGLM | 84.81 | 84.81 | 44.85 | 84.27 | 84.31 | 53.49 |

3) GCNet [25] (Graph Completion Network): Captures dependencies between time and speakers using speaker and temporal graph neural networks, jointly optimizing classification and reconstruction.
4) FDMER [26] (Feature-Disentangled Multimodal Emotion Recognition): Developed by Yang et al., learns shared and private feature representations for each emotion, exploring commonalities across different modalities.
5) MCTN [27] (Multimodal Cyclic Translation Network): Proposed by Pham et al., uses cycle consistency loss to enhance modality translation, ensuring maximal information retention from all modalities.
6) GraphCAGE [28] (Graph Capsule Aggregation) generates cross-modal feature representations through a cross-attention network when handling unaligned multimodal sequences. Simultaneously, it uses graph convolutional neural networks to learn dependencies between different time steps, ensuring information integrity and preventing information loss.
7) CORECT [29] (Conversational Context-Enhanced Temporal Graph Neural Network) employs a Relation-aware Temporal Graph Convolutional Network (RT-GCN) to learn local contextual representations by utilizing modality-level topological relationships. It infers global contextual representations across the entire conversation through pairwise cross-modal feature interactions (P-CM).
8) COGMEN [30] (Contextualized Graph Neural Network) uses graph neural networks to model complex dependencies in dialogues, capturing local information (i.e., interactions or internal dependencies between speakers) through the graph network. Nodes are viewed as queries, with neighboring nodes as global information (context), and node information is aggregated using a Graph Transformer.

The performance comparison on the CMU-MOSI dataset is summarized in Table II. Our method (MIAR) achieves superior results in terms of Acc2 and F1 metrics, outperforming IMDer and DiCMoR by 0.89%. However, it shows lower performance on Acc7 (40.67%), possibly due to limitations in dataset size for achieving the required representation in a multi-class setting. Overall, the experiments demonstrate that our method significantly enhances feature representations across modalities, effectively boosting performance in binary classification metrics.

The experimental results on the CMU-MOSEI dataset, shown in Table III, indicate that our method performs well, achieving good recognition rates in Acc2 and F1 metrics (86.27% and 86.23% respectively). It outperforms FDMER by 0.17% and 0.43% in these metrics but falls slightly behind in Acc7 (53.79%). Nevertheless, it surpasses IMDer and DiCMoR by 0.39%. These results collectively demonstrate that our approach significantly enhances feature representations across modalities, effectively improving emotion recognition performance.

*2) Results from Different Pretrained Language Models (Single Modality):* To validate the positive impact of features extracted by existing large language models on emotion recognition, this section presents results based on these models, as shown in Table IV.

From the table, it is observed that on the CMU-MOSI dataset, text features based on ChatGLM and CLAP (voice-text) models outperform those based on the BERT model. However, on the CMU-MOSEI dataset, CLAP's text encoder features underperform compared to BERT in binary classification and F1 scores. Across both datasets, ChatGLM's text features demonstrate the highest performance metrics.

These findings underscore the effectiveness of leveraging advanced language models for extracting textual features that enhance sentiment analysis across different multimodal datasets.

*3) Impact of Different Alignment Methods on Accuracy:* In this chapter, alignment methods proposed include norm alignment and contrastive learning. This subsection presents ablation experiments on these two components.

*a) Impact of Different Norm Alignment Methods:* Table V illustrates the results of different norm alignment methods on the CMU-MOSI dataset. The experiments indicate that norm alignment strategies have a subtle impact on the experimental metrics. Specifically, employing L1 norm alignment surpasses L2 norm alignment across all three metrics. Notably, it achieves a 0.31% increase in binary classification accuracy, highlighting the advantage of L1 norm in feature selection.

TABLE V
IMPACT OF DIFFERENT NORM ALIGNMENTS

| Norm Alignment | Acc2 (%) | F1 (%) | Acc7 (%) |
|---|---|---|---|
| L1 Norm | 86.59 | 86.57 | 40.67 |
| L2 Norm | 86.28 | 86.19 | 38.05 |

*b) Impact of Different Alignment Strategies on Accuracy:* Table VI presents the results of our proposed two alignment methods on the CMU-MOSI and CMU-MOSEI datasets.

Based on the previous experimental analysis, it was determined that L1 norm alignment performs better than L2 norm

TABLE VI
MODEL PERFORMANCE COMPARISON UNDER DIFFERENT ALIGNMENT STRATEGIES

| Dataset | Contrastive Alignment | Norm Alignment | Acc2 (%) | F1 (%) | Acc7 (%) |
|---|---|---|---|---|---|
| CMU-MOSI | ✗ | ✗ | 84.91 | 84.91 | 33.82 |
| | ✗ | ✓ | 85.06 | 85.15 | 37.90 |
| | ✓ | ✗ | 86.28 | 86.31 | 40.58 |
| | ✓ | ✓ | 86.59 | 86.57 | 40.67 |
| CMU-MOSEI | ✗ | ✗ | 84.07 | 84.14 | 49.27 |
| | ✗ | ✓ | 84.78 | 84.81 | 50.2 |
| | ✓ | ✗ | 85.94 | 86.04 | 52.96 |
| | ✓ | ✓ | 86.27 | 86.23 | 53.79 |

alignment in this scenario. Subsequently, experiments were conducted using L1 norm alignment.

For the CMU-MOSI dataset, compared to the scenario without alignment strategies, integrating contrastive alignment and norm alignment methods separately improves the accuracy of seven-class classification by 6.78% and 4.08%, respectively. This indicates significant effects of these two methods on multi-class accuracy, verifying the subtle differences learned in positive samples' features that influence multi-class accuracy. After adding both strategies, our method outperforms in all three metrics by 1.68%, 1.66%, and 6.85%, respectively.

Similarly, on the CMU-MOSEI dataset, compared to the absence of alignment strategies, integrating contrastive alignment and norm alignment methods separately improves the accuracy of seven-class classification by 3.68% and 0.93%, respectively. Similar to the results on the CMU-MOSI dataset, these two strategies promote multi-class tasks, but the improvement is not as significant as the former. This may be attributed to the larger number of samples, allowing larger batch sizes for training models, enabling the model without alignment strategies to learn better model parameters. After adding both strategies, our method outperforms in all three metrics by 2.2%, 2.09%, and 4.52%, respectively.

The performance across these three metrics on both datasets demonstrates the effectiveness of our approach in emotion prediction tasks.

The table compares the performance of the model under different alignment strategies. Here's the translation of the table caption and the content:

*4) Mapping Dimension of Alignment Space:* This section explores the impact of mapping dimension $d$ to the alignment space on the results. The cross-modal interaction fusion module in this paper has a dimension of 50. Alignment space dimensions are set to 16, 32, 64, and 128, and ablation experiments are conducted on the CMU-MOSI dataset, as shown in Table VII. Among the four dimensions set, a dimension of 32 achieves the highest binary classification accuracy and F1 score, increasing by 0.92% and 0.91%, respectively, compared to the projection dimension of 64. Additionally, it achieves a higher seven-class accuracy of 0.18% compared to the 64-dimensional case. These results indicate that appropriately reducing the feature dimensions can help the model learn more general features and patterns, leading to better generalization to unseen data.

These findings underscore the importance of selecting an appropriate alignment space dimension $d$ to optimize the model's performance in multimodal emotion recognition tasks.

TABLE VII
IMPACT OF ALIGNMENT SPACE DIMENSION $d$ ON RESULTS

| Alignment Space Dimension | Acc2 (%) | F1 (%) | Acc7 (%) |
|---|---|---|---|
| 16 | 84.45 | 84.41 | 37.46 |
| 32 | 86.59 | 86.57 | 40.67 |
| 64 | 85.67 | 85.66 | 40.49 |
| 128 | 85.52 | 85.58 | 38.48 |

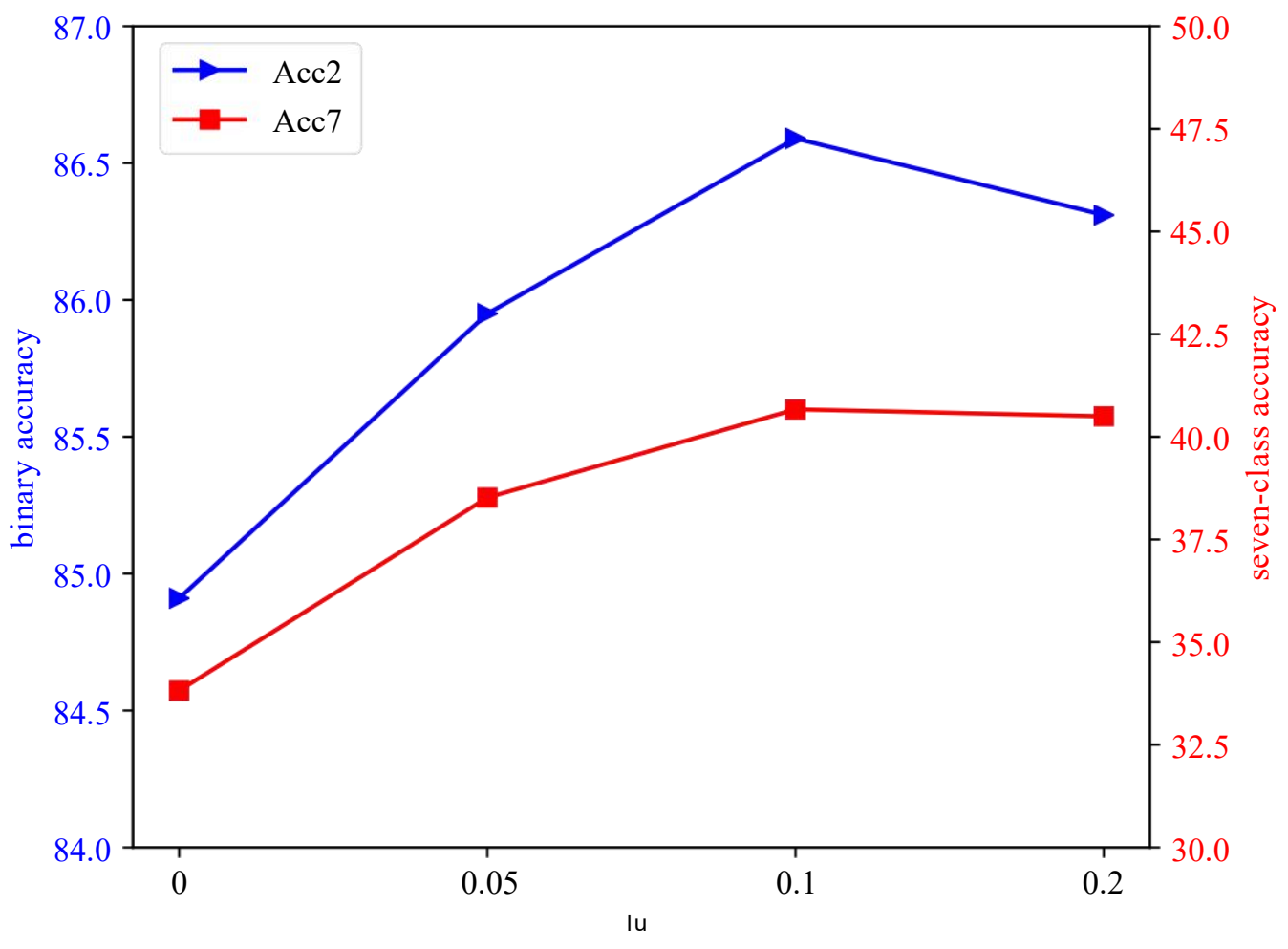

Fig. 6. Performance of different loss weight $\omega$ on CMU-MOSI

*5) Impact of alignment loss weight $W$.:* During the training of the network model, the loss function defined in 16 will be optimized. The loss weight $W$ represents the relative importance of the regression loss and the alignment loss. This section explores the impact of different $W$ values on the performance of the network model on CMU-MOSI. Figure 5.4 shows the recognition accuracy when the parameter $W$ changes from 0 to 0.2. Note that $W = 0$ means that only the regression loss is used for optimization. From Figure 6, it can be seen that as $W$ gradually increases, the recognition accuracy also gradually improves, but a further increase leads to a decrease in recognition accuracy. From the overall results, it can be seen that the alignment loss affects the performance of multimodal emotion recognition. Therefore, $W = 0.1$ is chosen as the overall loss weight for the final optimization function.

## V. Conclusion

The paper introduces challenges posed by the complementarity and distribution differences of heterogeneous modalities in multimodal settings. It proposes a Cross-Modal Interaction Alignment Representation Fusion Prediction Network, comprising a feature interaction module and a temporal enhancement module. The feature interaction module integrates contextual features across different modalities, generating feature tokens representing global representations of how each modality extracts information from others. These representations are then projected into an alignment space. The chapter also discusses modality alignment strategies, including contrastive learning and norm alignment (L1 and L2). Experimental validation on two emotion datasets confirms the effectiveness of the proposed methods, with additional ablation experiments analyzing the impact of different components on emotion recognition.

## Acknowledgments

This work was supported by the Natural Science Foundation of China (62276242), National Aviation Science Foundation (2022Z071078001), CAAI-Huawei MindSpore Open Fund (CAAIXSJLJJ-2021-016B, CAAIXSJLJJ-2022-001A), Anhui Province Key Research and Development Program (202104a05020007), USTC-IAT Application Sci. & Tech. Achievement Cultivation Program (JL06521001Y), Sci. & Tech. Innovation Special Zone (20-163-14-LZ-001-004-01).

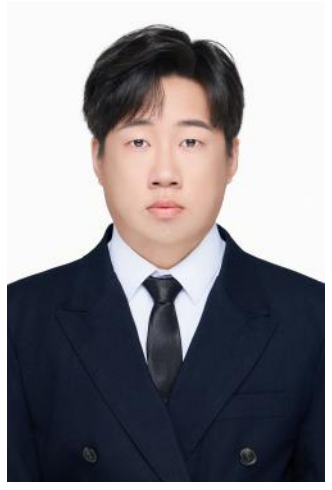

**Jichao Zhu** received the B.S. degree from Chang'an University, Xi'an, China, in 2021, and he is currently working toward the M.S. degree with the Department of Automation, University of Science and Technology of China, Hefei, China. His current research interests include affective computing, deep learning, and multimodal representation learning.

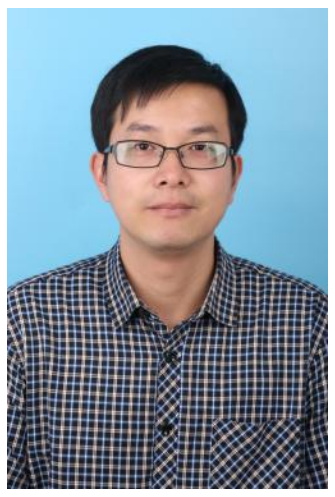

**Jun Yu** is currently an associate professor and laboratory director with the Department of Automation and the Institute of Advanced Technology, University of Science and Technology of China. His research interests are Multimedia Computing and Intelligent Robot. He has published 200+ journal articles and conference papers in TPAMI, IJCV, JMLR, TIP, TMM, TASLP, TCYB, TITS, TCSVT, TOMM, TCDS, ACL, CVPR, ICCV, NeurIPS, ICML, ICLR, MM, SIGGRAPH, VR, AAAI, IJCAI, etc. He has received 6 Best Paper Awards from premier conferences, including CVPR PBVS, ICCV MFR, ICME, FG, and won 60+ champions from Grand Challenges held in NeurIPS, CVPR, ICCV, MM, ECCV, IJCAI, AAAI. Email: harryjun@ustc.edu.cn.